\documentclass[journal]{IEEEtran}                   

\usepackage{booktabs}
\usepackage{cite}
\usepackage[font=footnotesize]{subfig}
\usepackage{multirow}
\usepackage{ifpdf}
\usepackage[pdftex]{graphicx}
\usepackage{epstopdf}
\usepackage[cmex10]{amsmath}
\usepackage{tabularx, booktabs, siunitx}

\usepackage{array}
\usepackage{mdwmath}
\usepackage{mdwtab}
\usepackage{eqparbox}
\usepackage{verbatim}

\usepackage{amsmath}
\usepackage{amssymb}
\usepackage{algorithm}

\usepackage{algorithmicx}
\usepackage{algpseudocode}

\usepackage[font=footnotesize]{subfig}
\usepackage{fixltx2e}
\usepackage{stfloats}

\usepackage{amsmath,bm}
\usepackage{amssymb}
\usepackage{color}
\usepackage{subfig}

\usepackage{multirow}
\usepackage{float}
\usepackage{caption}

\usepackage{comment}

\usepackage{makecell}
\usepackage{amsmath}
\usepackage{graphicx}
\usepackage[colorlinks=true, allcolors=blue]{hyperref}
\usepackage{adjustbox}
\usepackage{blindtext}
\usepackage{hyperref}
\usepackage{tabularx} 
\usepackage[font=footnotesize]{subfig}
\usepackage{fixltx2e}
\usepackage{stfloats}

\usepackage{amsmath,bm}
\usepackage{amssymb}
\usepackage{color}
\usepackage{subfig}

\usepackage{multirow}
\usepackage{float}
\usepackage{caption}
\usepackage[normalem]{ulem}
\usepackage{comment}
\usepackage{hyperref}
\usepackage{cite}
\usepackage[font=footnotesize]{subfig}

\usepackage{ifpdf}

\usepackage{epstopdf}
\usepackage[cmex10]{amsmath}

\usepackage{array}
\usepackage{mdwmath}
\usepackage{mdwtab}
\usepackage{eqparbox}
\usepackage{verbatim}
\usepackage{booktabs}
\usepackage{multirow}
\usepackage{graphicx}

\usepackage{array} 

\definecolor{orange}{rgb}{1,0.5,0}

\begin{document}
	
\title{\LARGE \bf
TacFR-Gripper: A Reconfigurable Fin Ray-Based Compliant Robotic Gripper with Tactile Skin for In-Hand Manipulation
}


\author{
Qingzheng Cong, Wen Fan, Dandan Zhang

\thanks{Q.Cong, W. Fan are with the Department of Engineering Mathematics, University of Bristol, affiliated with Bristol Robotics Lab. 
D. Zhang is with the Department of Bioengineering, Imperial College London, and the I-X initiative. }
}
\maketitle

\begin{abstract}
This paper introduces the TacFR-Gripper, a reconfigurable Fin Ray-based soft and compliant robotic gripper equipped with tactile skin, which can be used for dexterous in-hand manipulation tasks. This gripper can adaptively grasp objects of diverse shapes and stiffness levels. An array of Force Sensitive Resistor (FSR) sensors is embedded within the robotic finger to serve as the tactile skin, enabling the robot to perceive contact information during manipulation. We provide theoretical analysis for gripper design, including kinematic analysis, workspace analysis, and finite element analysis to identify the relationship between the gripper's load and its deformation. Moreover, we implemented a Graph Neural Network (GNN)-based tactile perception approach to enable reliable grasping without accidental slip or excessive force.

Three physical experiments were conducted to quantify the performance of the TacFR-Gripper. These experiments aimed to i) assess the grasp success rate across various everyday objects through different configurations, ii) verify the effectiveness of tactile skin with the GNN algorithm in grasping, iii) evaluate the gripper's in-hand manipulation capabilities for object pose control. 
The experimental results indicate that the TacFR-Gripper can grasp a wide range of complex-shaped objects with a high success rate and deliver dexterous in-hand manipulation. Additionally, the integration of tactile skin with the GNN algorithm enhances grasp stability by incorporating tactile feedback during manipulations.
For more details of this project, please view our website: \url{https://sites.google.com/view/tacfr-gripper/homepage}. 
\end{abstract}

\section{Introduction}
In recent years, soft and compliant robotic grippers have transformed the landscape of robotics by offering unparalleled flexibility, adaptability, and safety compared to traditional rigid grippers \cite{alves2023integrated}. A typical example of this innovation is the Fin Ray-inspired robotic gripper, which draws inspiration from the fin movements of fish to deliver a versatile and effective gripping mechanism \cite{9981987}. It has high flexibility and can easily adapt to an object's contour, thus having a greater contact area for grasping \cite{dong_bozkurttas_mittal_madden_lauder_2010, 8813388,https://doi.org/10.1002/adma.201707035}.
However, most of the existing Fin Ray-inspired soft robotic grippers do not have in-hand manipulation functions, which lack the ability to reposition objects within a robotic hand through the coordination of relative motions between the object and the hand \cite{bullock_classifying_2011,crooks2016fin,lu_systematic_2021}.  Therefore, our research aims to design a reconfigurable Fin Ray-based robotic gripper, specifically for in-hand manipulation.

Another important factor in in-hand manipulation is tactile perception \cite{in1}, as it provides essential information about the object's pose, shape, stiffness and other characteristics, thereby facilitating reliable manipulation. However, there still remains a research gap in integrating tactile feedback mechanisms with soft robotic grippers to enhance their reliability and dexterity for in-hand manipulation of complex-shaped objects.
It is important to augment the soft and compliant gripper with integrated tactile sensing capabilities to facilitate intricate object manipulations \cite{gelsight}. To this end, we develop a soft robotic gripper with in-hand manipulation as well as tactile sensing
 capabilities.

\begin{figure}[t]
    \centering
\captionsetup{font=footnotesize,labelsep=period}
    \includegraphics[width = 0.95\hsize]{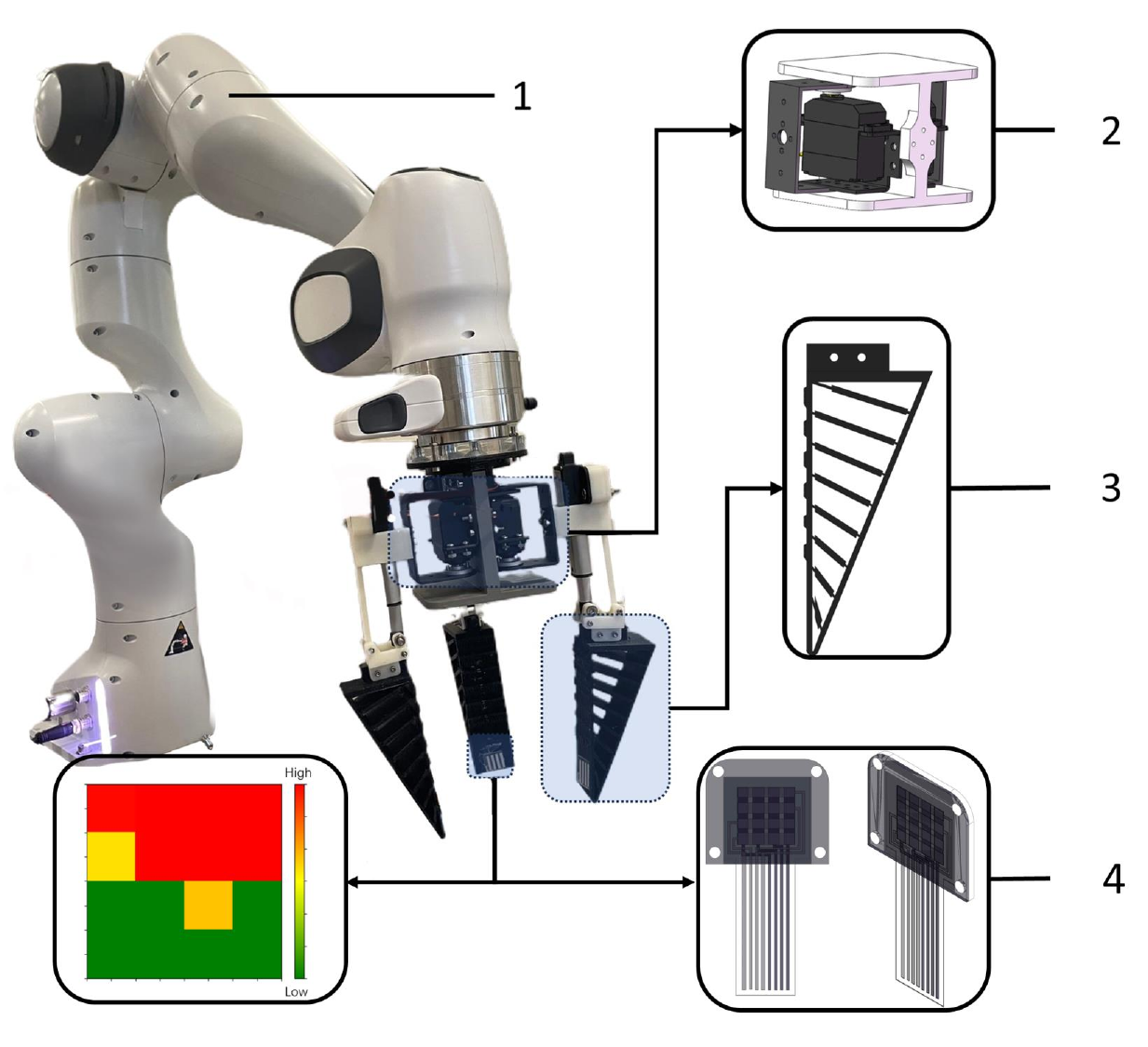}
    \caption{Overview design of the soft gripper with tactile feedback. (1-Franka Emika Robot. 2-Reconfigurable Gripper Base driven by servo motors. 3-Fin Ray-inspired Finger. 4-Tactile skin and visualization of the tactile image obtained by the embedded sensor in the tactile skin. }
    \label{overview}
    \vspace{-0.6cm}
\end{figure}

Our research seeks to bridge this gap with the introduction of the TacFR-Gripper, a reconfigurable Fin Ray-inspired soft robotic gripper equipped with tactile skins. This paper details the design, development, and testing processes of the TacFR-Gripper. We introduce the integration of the Fin Ray-inspired soft finger and the tactile skin, which mimics the properties of human skin by being able to measure external forces and subsequently adjust the pose of the fingers to accommodate the object in hand. The tactile skin on each fingertip comprises an array of sensing elements, which allow the robot to perceive contact information during manipulation and adjust its grip accordingly. Furthermore, our approach employs machine learning-based tactile perception to facilitate reliable grasping, aiming to significantly enhance tactile perception capabilities. More specifically, we use Graph Neural Networks (GNN) for interpretable and data-efficient tactile sensing. This is due to the fact that tactile data often possess highly irregular structures, making them well-suited for representation as graphs \cite{10160288}.


In summary, the main contributions of this paper are listed as follows:
\begin{itemize}
    \item Development of a low-cost, bio-inspired, soft, and compliant robotic gripper with in-hand manipulation capabilities.
    \item Integration of the aforementioned robotic gripper design with tactile skin to enhance the perception of contact force.
    \item Utilization of a GNN-based method for reliable tactile perception during robotic manipulation.
\end{itemize}



\section{Related Work}

\subsection{Soft Robotic Gripper with Tactile Perception}

Tactile sensing in soft robotic grippers can help improve object manipulation reliability by providing direct contact information, which is crucial for precise in-hand manipulation and feedback control. 
 Sakuma et al. developed a granular-jamming-based universal soft gripper \cite{sakuma2018universal} that was filled with transparent objects. This gripper included an internal camera to trace markers on its surface, enabling a 3D reconstruction of the contact surface.
A soft gripper named TaTa \cite{TATA} incorporates RGB tricolor LEDs and employs an inbuilt camera to monitor color changes on the contact surface, significantly enhancing the resolution of the tactile image. However, integrating rigid elements such as camera modules, circuit boards, and lights into soft fingers without interfering with the soft gripper's active deformation poses a challenge.

Zhao et al. introduced GelSight Svelte \cite{zhao2023gelsight}, a soft, finger-sized device that employs a camera to capture the reflection from a curved mirror, providing proprioceptive and tactile information. However, the GelSight Svelte design is limited due to its lack of compliant properties.
Sandra et al. equipped a Fin Ray-inspired finger with the GelSight tactile sensor \cite{yuan2017gelsight}, enabling the detection of an object's pose and facilitating tactile reconstruction. This setup proved effective for reorienting a glass jar \cite{gelsight}. However, the delicate nature of the GelSight sensor, tightly integrated with the finger's structure, poses durability concerns. Wear and tear, such as tearing at the skin's connection to the housing or elastomer shearing from intensive manipulation or collisions, can damage the sensor. These issues potentially limit the gripper's service life and reduce its cost-effectiveness


To acquire comprehensive contact information while ensuring the affordability of the gripper, we aim to integrate a flexible tactile sensor array onto the surface of the robotic finger as tactile skin.
Although tactile sensor arrays may exhibit drawbacks like relatively low resolution compared to vision-based tactile sensors \cite{he2023tacmms,li2022implementing}, they have inherent advantages in terms of flexibility and adaptability. Their design facilitates seamless integration into robots, and their rapid response time makes them exceptionally suited for in-hand manipulation tasks.

\subsection{Reconfigurable Soft Gripper for In-Hand Manipulation}

In recent years, reconfigurable soft grippers have gained increasing attention in robotics, since they can adjust their shapes and functions based on task needs. This adaptability allows for in-hand manipulation, which can significantly enhance the versatility of the robotic gripper \cite{zhakypov2018origami}.

A versatile soft robotic gripper system \cite{Low} has been developed by Jin et al., which allows the fingers to be reconfigured into various poses, including scoop, pinch, and claw, utilizing the base of the hand. Since each finger operates independently in this design, it has advantages not only in grasping diverse objects but also in performing intricate in-hand manipulations. 
Batsuren et al. focused on a gripper with three pneumatic fingers \cite{Batsuren_2019}, each containing three air chambers: two for twisting in different directions and one for grasping. This design enables the gripper to handle objects of varying shapes and sizes.
This design was later improved by inserting a stiff rod that can be maneuvered inside the central hole of each finger to control the bending point \cite{Pagoli}. This new configuration significantly enhances the dexterous grasping capabilities of the soft gripper, with a specific emphasis on in-hand manipulations. Jain et al. developed a gripper equipped with retractable fingernails and a versatile palm \cite{nails}, thereby enhancing its capability for various manipulation tasks. This innovative design enables the gripper to exert up to 1.8N in normal grasping forces and to handle objects as thin as 200$\mu$m on flat surfaces.



However, the grippers previously mentioned require molding processes that are time-consuming and prone to manufacturing errors. Furthermore, the functionality of these soft grippers heavily depends on the deformation of their silicone skin, which can lead to operational inaccuracies. Therefore, we explore the use of multi-material 3D printing technology to fabricate entire soft grippers efficiently and cost-effectively.  With this approach, we aim to mitigate the manufacturing and operational limitations of traditional soft grippers and facilitate precise object handling and manipulation.



\section{Methodology}
\subsection{Hardware Design}
\subsubsection{Sensor Array}
The tactile skin on the robotic gripper comprises two distinct layers, as shown in Fig. \ref{overview} (4). The external layer is fabricated using a multi-material 3D printing technique, utilizing a rubbery polymer known as Agilus30. The internal layer, crucial for the sensor's functionality, consists of a Force Sensitive Resistor (FSR) array. This FSR array is designed to emulate human tactile perception by incorporating a dense array of mechanoreceptors. 
The sensor array employs nanomaterials that can alter their electrical properties under slight forces, transforming physical pressure into measurable electrical signals. Moreover, silver paste is integrated within the sensor for its superior electrical conductivity \cite{duan2022recent}, which facilitates the development of flexible and robust circuits that are crucial for accurately detecting force within the sensor array. Each sensor unit functions as a variable resistor, which 
 can dynamically alter its resistance based on the applied force, thereby enabling precise distributed force map measurement.

\subsubsection{Robot Gripper Design}

\begin{figure*}[t]
\centering
\captionsetup{font=footnotesize,labelsep=period}
\includegraphics[width=0.95\textwidth,height=3.9in]{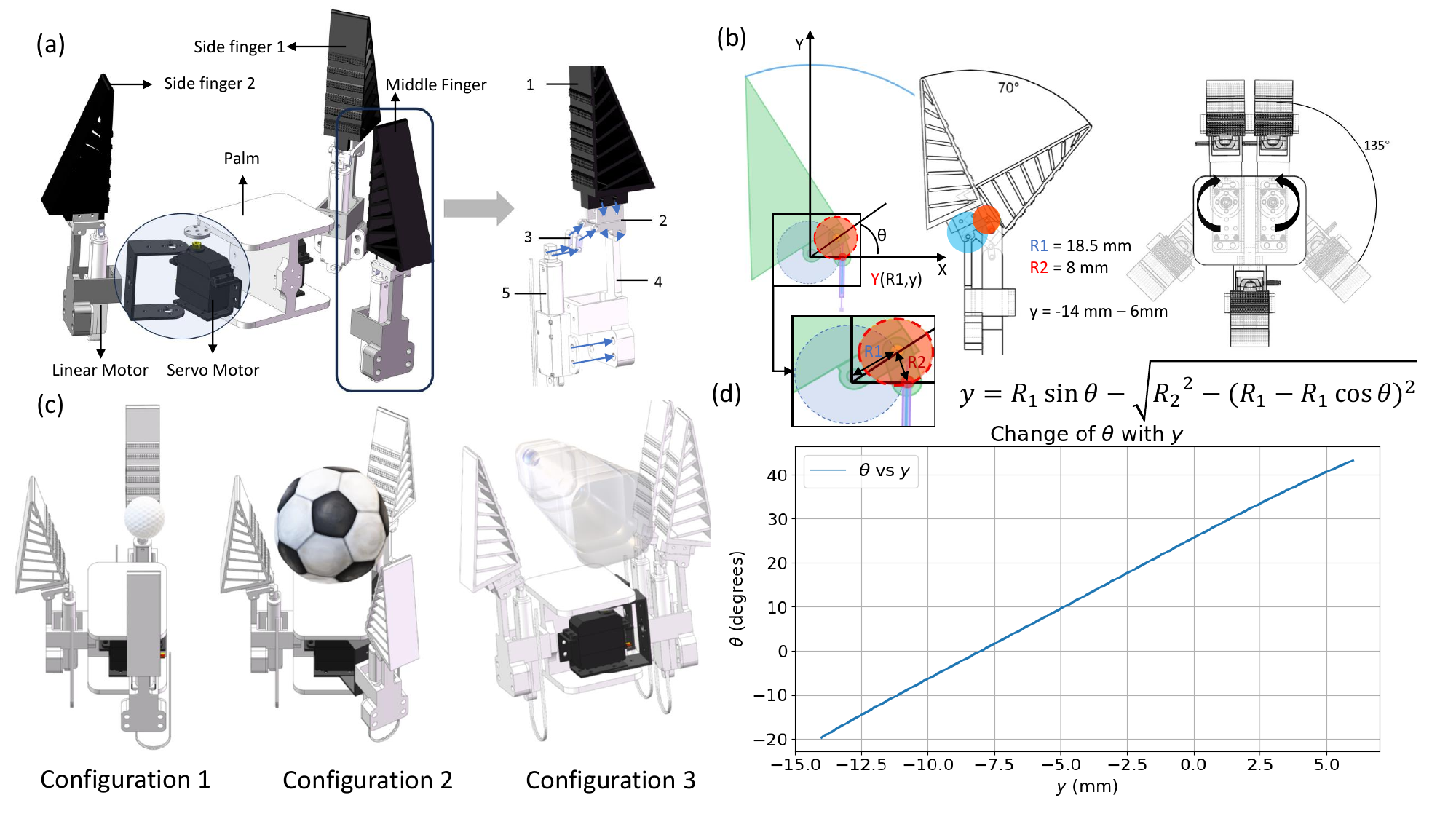}
    \caption{Hardware overview. (a) Exploded view of the robotic gripper, 3D-printed components (2, 3, 4), a four-bar mechanism, and a linear actuator (5). (b) Kinematic diagram of the finger's range of motion, with R1 and R2 denoting pivot radii, and 'y' representing the linear actuator's endpoint displacement. $\theta$ indicates the finger's bend angle. (c) Illustrative configurations of the gripper that demonstrate parallel, trigonal, and T-shaped grasps. (d) A graph of the actuator's y-displacement against finger bend angle $\theta$ based on mathematical modeling.}
\label{hardware}
\vspace{-2mm}
\end{figure*}

Fig. \ref{hardware} (a) illustrates the mechanical structure of the robotic gripper, while Fig. \ref{hardware} (b) displays the kinematic diagram of the finger and its motion range. The gripper consists of four main components: a Fin Ray-inspired robotic finger, a four-bar mechanism, a linear actuator for controlling finger bending motion, and a palm equipped with two servo motors to reposition the side fingers relative to the palm.
There is an approximately linear relationship between the coordinates of the linear actuators and the angular displacement of the finger in the four-bar linkage mechanism. This correlation is further elucidated in Fig. \ref{hardware} (d), where the relationship between the finger’s bending angle $\theta$, and the displacement of the linear actuator $y$ is analyzed through mathematical modeling.

The reconfiguration of the gripper is enabled by two servo motors assembled in the hollow area of the palm, which grant rotational movement to the side fingers to enable reconfigurability of the gripper. Thus, the robotic gripper can adjust its grips for different types of objects.
Fig. \ref{hardware} (c) illustrates three general configurations of the gripper. Configuration 1 engages solely in using the side fingers to grasp targeted objects. Configuration 2 involves the collaborative effort of all three fingers, making it suitable for grasping objects with spherical shapes. Configuration 3 is used for grasping elongated objects in a horizontal position. These configurations exemplify the robotic gripper's adaptive design and its ability to cater to a wide range of manipulative functions in robotic applications.


\begin{figure}[!htbp]
            \centering
\captionsetup{font=footnotesize,labelsep=period}    \includegraphics[width=0.45\textwidth,height = 1.3in]{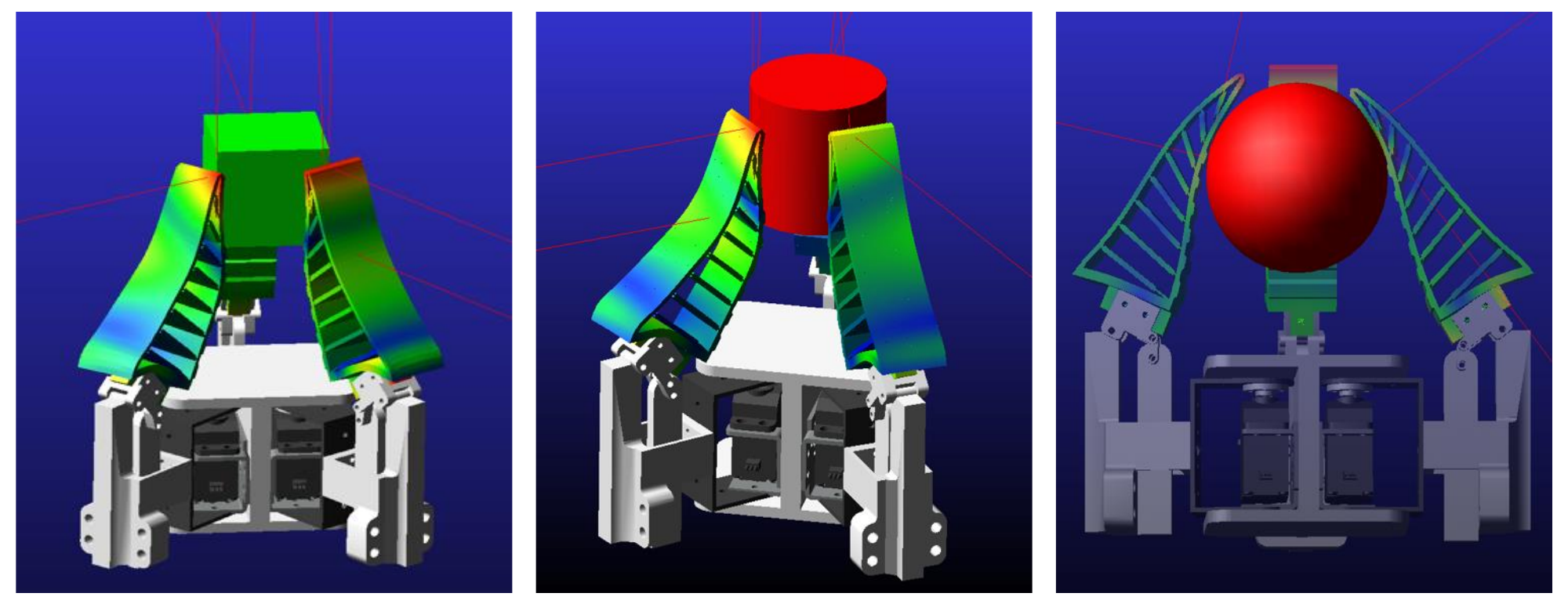}
              \caption{Visualization of the maximum load analysis using ADAMS simulation when grasping objects with different shapes. }
              \label{fig:combined}
\end{figure}

\begin{figure}[!htbp]
            \centering
\captionsetup{font=footnotesize,labelsep=period}    \includegraphics[width=0.45\textwidth,height = 1.8in]{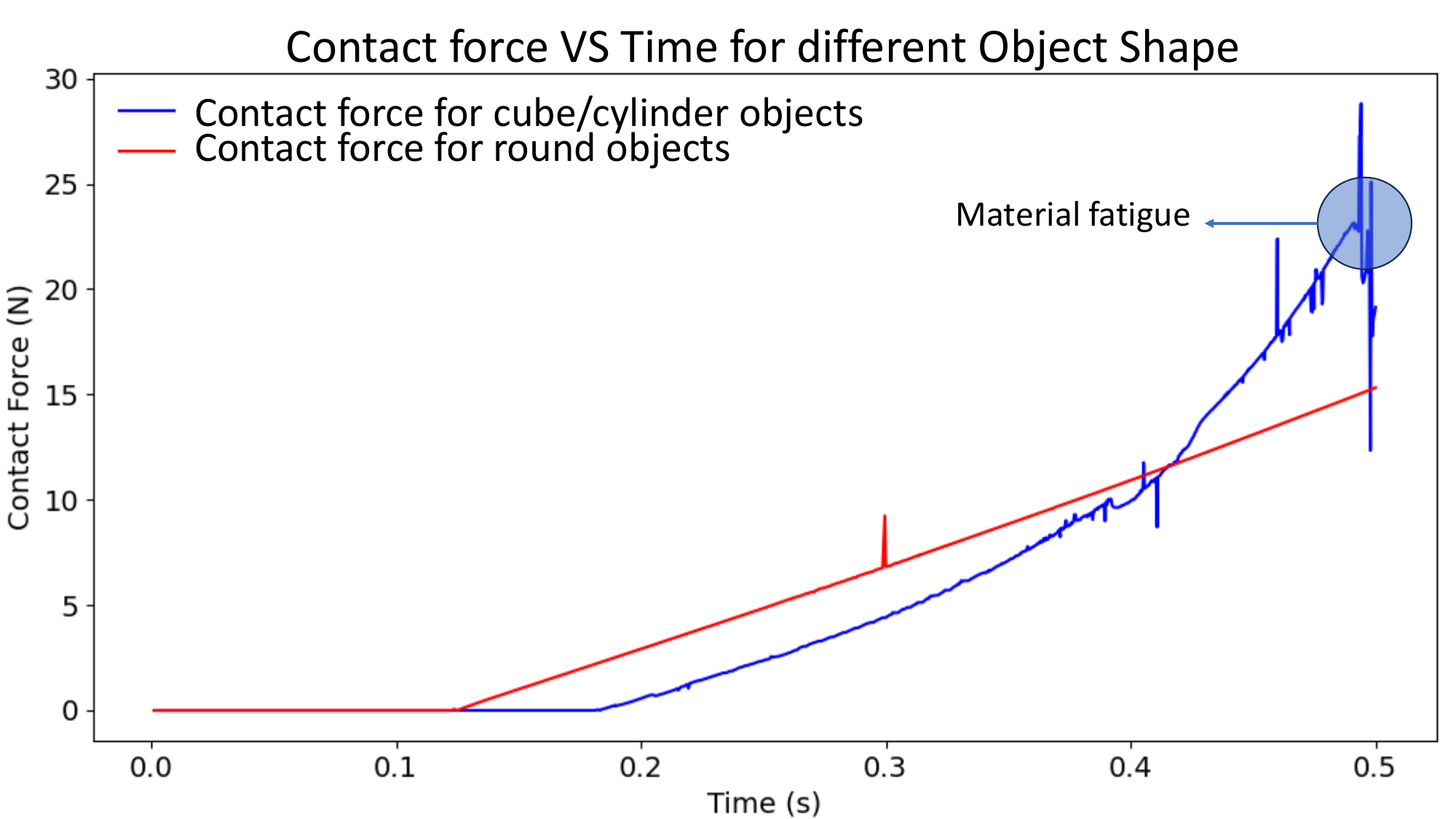}
              \caption{Contact force for grasping objects with different shapes, including cube, cylinder, and round objects. }
              \label{fig:gesture 12c}
              \vspace{-3mm}
\end{figure}

\begin{figure*}[t]
    \centering
\captionsetup{font=footnotesize,labelsep=period}
    \includegraphics[width=1\textwidth, height = 3.7 in]{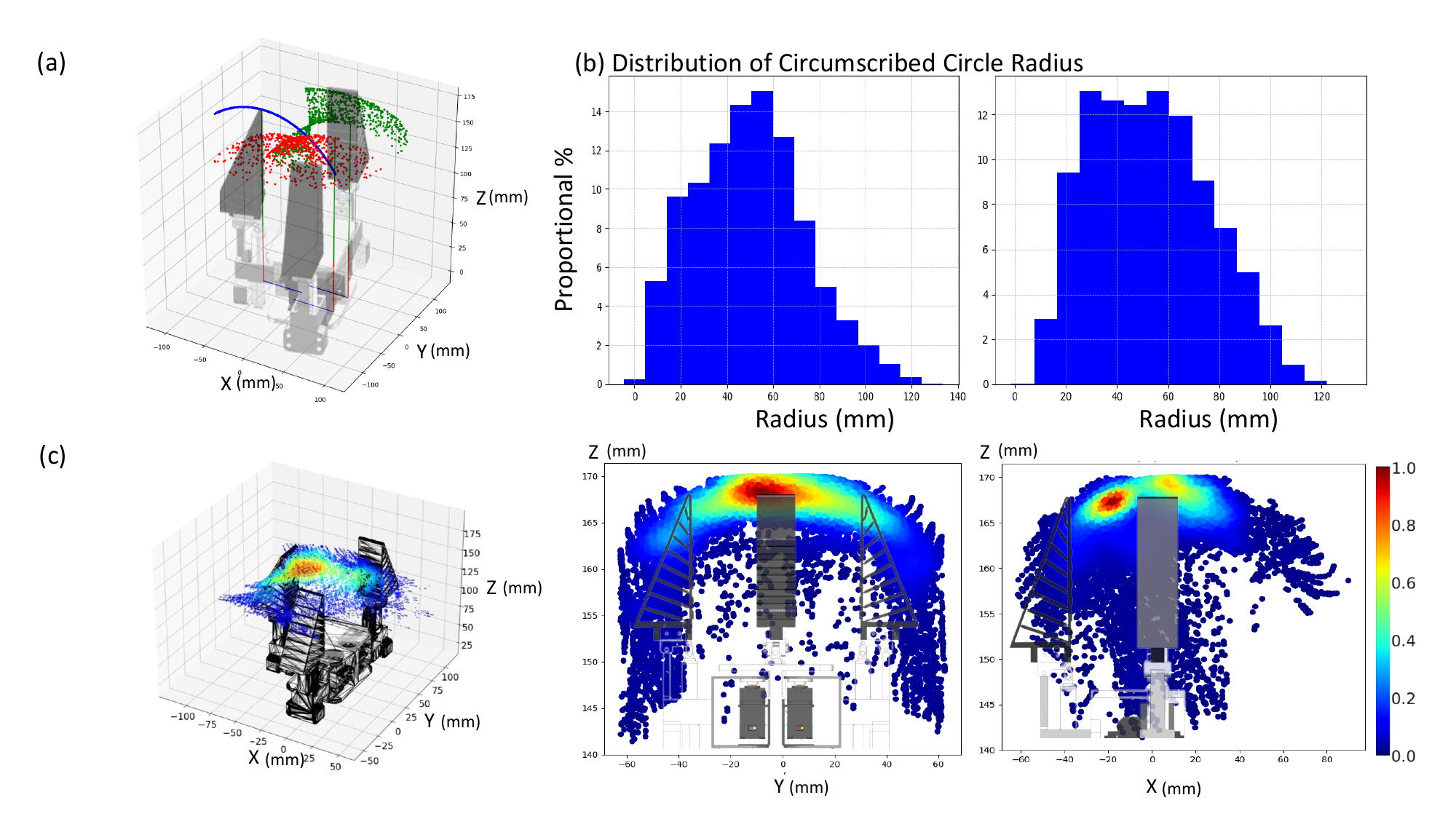}
    \caption{(a) Visualization of the reachable workspace analysis for gripper based on Monte Carlo simulation. (b) Histograms showing the distribution of circumcircle radius for Configuration 1 (left) and Configuration 2, 3 (right), indicating the overall trend of graspable object sizes. (c) 3D view of the dexterity workspace and its cross-sectional views along the X and Y axes for objects with radius between 60 and 80 mm.}
    \label{fig_sim}
    \vspace{-3mm}
\end{figure*}

\subsection{Theoretical Analysis}

\subsubsection{Force Analysis}

In the maximum load analysis, simulation is carried out in ADAMS based on the principles of theoretical mechanics and multi-body dynamics. In the simulation, an object is positioned at the center of the robotic fingers to evaluate grasping capabilities. Cubes, cylinders, and round objects are used for simulation, as shown in Fig. \ref{fig:combined}. 

During the simulation with the cylinder, the maximum deformation observed in the gripper is 13.62 mm. In a similar test featuring the cube, the deformation is marginally less, measuring 12.89 mm. When the tip of a Fin-Ray structure is excessively pressed, it exhibits a pronounced outward deformation in the opposite direction. This effect is particularly noticeable with shapes vastly different from a round object such as a sphere. Given its continuous surface in all directions of the sphere, the Fin-Ray structure can grip more tightly without creating gaps or losing contact. This leads to a larger contact area between the object and the gripper \cite{agriculture12111802}, resulting in a significantly smaller deformation, measured at 7.6 mm.

Fig. \ref{fig:gesture 12c} displays the gripper's maximum force over time. For the cube and cylinder, their maximum forces are around 25 N, while for the sphere, it is 12 N. This result closely aligns with the deformation observed. Additionally, since the contact area for a spherical object is generally larger than that of the cube and cylindrical ones, the maximum pressure under the same force for the sphere should also be much smaller.

\subsubsection{Workspace Analysis}
Workspace analysis is crucial in robotics for understanding the capabilities of robotic manipulators or grippers for manipulation \cite{zhang2020ergonomic,workspacet,zhang2019design}.
Here, we use the Monte Carlo method to simulate a large number of potential locations for the fingers' tip \cite{Montec,zhang2019wsrender}, taking into account the physical constraints imposed by the joints.
Fig. \ref{fig_sim} (a) displays the potential contact points between the gripper's three fingers and the targeted object for grasping. We calculate the circumradius of triangles formed by these contact points to construct a histogram. This histogram contrasts the gripper's graspability for specific dimensions (radius) among its total grasping range, showcasing the gripper's effectiveness with various object sizes.
In Fig. \ref{fig_sim} (b), the left side illustrates Configuration 1, tailored for grasping smaller objects using a pinching method. The right side of the same figure represents Configurations 2 and 3, which demonstrates the gripper's suitability for handling medium-sized objects, specifically those with a radius between 20 to 80mm.

For visualization, Kernel Density Estimation (KDE) with a Gaussian Kernel is used to create smooth, continuous density estimations\cite{scipy_gaussian_kde}. We apply KDE to the workspace visualization of the proposed gripper, which demonstrates where the gripper is most capable of handling objects of various sizes. In Fig. \ref{fig_sim} (c), the workspace analysis and visualization of the gripper for grasping objects with varying radii (from 60 to 80mm) is shown. Darker shades like blue indicate lower accessibility, while brighter regions (reds and yellows) indicate higher dexterity. 

However, not all dexterity workspace is continuous like Fig. \ref{fig_sim}. When manipulating objects with a radius of 80-100 mm, the gripper's workspace features two highly dexterous regions on both sides. However, these regions are separated by an area with lower dexterity, indicating a gap in the dexterity workspace for objects transitioning between them. For larger objects, specifically those with a radius between 100-120 mm, the workspace lacks such a gap but is rather confined and concentrated within a limited area. This limitation results in a significant reduction in the gripper's maneuverability when handling objects that vary in size. The kinematic structure of the reconfigurable gripper can be further enhanced through optimization.

\subsection{Tactile Perception For Slip Detection}



\begin{figure}[t]
            \centering
\captionsetup{font=footnotesize,labelsep=period}            \includegraphics[width=0.45\textwidth,height = 2in]{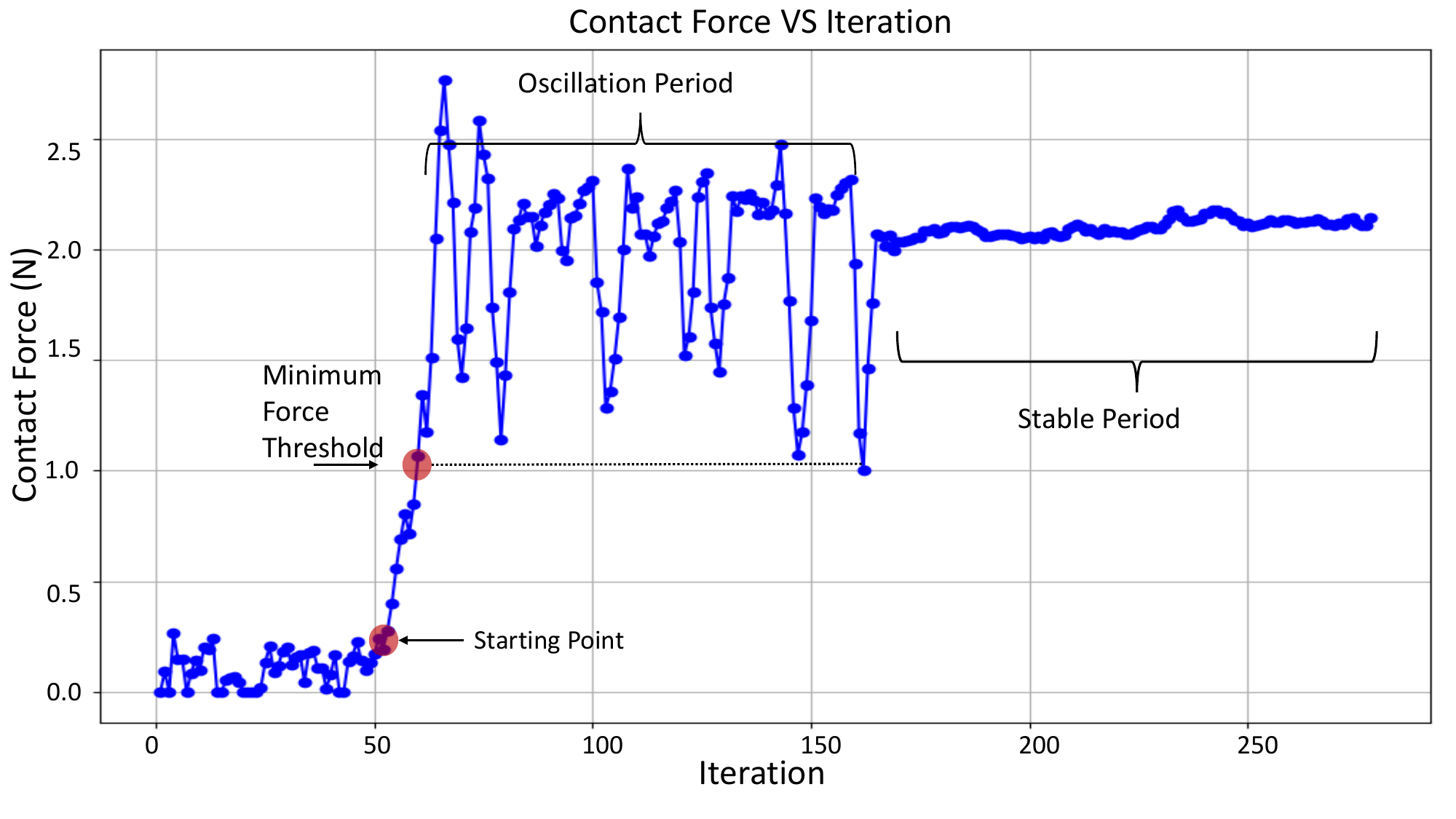}
\caption{Tactile sensor readings and contact force recorded from a grip test: Illustration of the stable period, oscillation period during grasping, identification of the grasping starting point, and the minimum force threshold.}
              \label{TactileReading}
\vspace{-3mm}              
\end{figure}

\subsubsection{Threshold-Based Method}
To determine the minimum force to securely grip objects without slipping, a calibration method of gradual reduction in gripping force is employed. Fig. \ref{TactileReading} shows the curve of the contact force during an object grasping process. We can easily identify the initial sensor noises, the oscillation period of unstable grasping, and the stable period of grasping. We can calculate the noise level when the gripper hasn't started to grasp objects. After filtering sensor noise, we can obtain a reasonable threshold value by calculating the mean of the remaining non-zero readings. We can infer that the lowest reading during this oscillation represents the minimum force threshold.
However, threshold-based methods require manual tuning of thresholds to adapt to different conditions. Therefore, we explore machine learning-based approaches, which can adapt to new operating conditions without re-tuning. 

\begin{figure}[t]
 \centering
\captionsetup{font=footnotesize,labelsep=period} \includegraphics[width=0.4\textwidth,height = 1.4in]{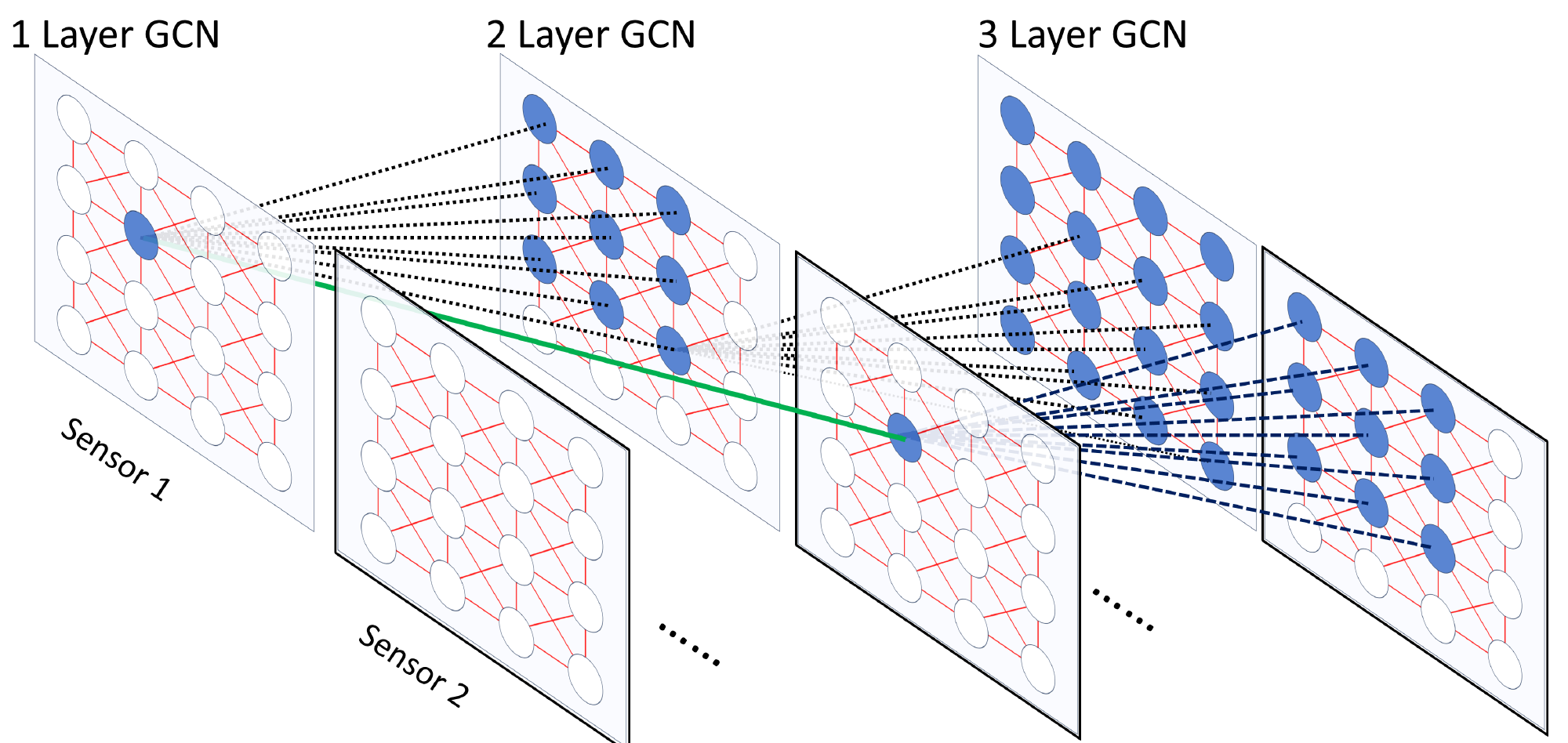}
\caption{GCN Architectures with 3 layers for sensor data integration. Blue dots represent sensor nodes, connected within sensor arrays by red lines for local feature extraction, and across arrays by green lines for data integration.}
\label{GCNS}
\vspace{-6mm} 
\end{figure}
            
\subsubsection{GNN-based Method}



GNNs excel at learning complex and nonlinear relationships in non-Euclidean \cite{fanwen}. This ability makes GNN models more accurate at detecting slip events compared to simplistic thresholding methods that rely on fixed rules. Additionally, they can infer the slip state smoothly even when encountering different objects without any manual tuning of threshold parameters.

Here, we model data from tactile sensors as a graph, where each sensor unit's value is represented as a node. These nodes are connected to their neighboring nodes within the same array.  A 4x4 sensor array is used as an example, while 84 edges are formed within each sensor array. We use a Graph Convolutional Network (GCN) in this paper, which is a type of GNN that applies convolutional operations directly to graphs. This approach learns features by inspecting neighboring nodes and aggregating node vectors through convolutional layer \cite{zhu2020simple}. The inner-connections within each layer are represented by dotted lines, as shown in Fig. \ref{GCNS}.  
 Each node is additionally connected to its corresponding nodes in the other two sensor arrays, effectively integrating data across three sensors and adding 32 edges per sensor. These inter-array connections are highlighted with blue lines. In total, the network comprises 348 edges, enabling the GCN to aggregate features from multiple layers. This process facilitates the analysis and extraction of intricate spatial relationships among the sensor arrays. By aggregating information from each node's immediate neighborhood in the first layer, and incrementally broadening this scope in subsequent layers, the GCN significantly enhances its capability to identify complex patterns across the three distinct sensor data sets.

The effectiveness of feature learning is highly dependent on the comprehensiveness of the training data. Thus, we collected a dataset from experiments using a threshold-based method, categorizing samples under the minimum force threshold as 0 (non-slip) and those above as 1 (slip), as depicted in Fig. \ref{TactileReading}. 
The dataset encompasses 2100 training and 700 test samples.  In this study, a three-layer GCN model trained on this dataset could achieve a high test accuracy of 96.2\%.

\section{Experiments}
\subsection{Gripping Capability Evaluation}

The first experiment aims to assess the gripping capability of the TacFR-Gripper. In this experiment, the gripper engages with objects of varying sizes and repeatedly grasps them ten times to evaluate the success rate across different configurations. The properties of the objects used in the gripping experiments are summarized in Table \ref{ex}. We evaluate the success rate by grasping different objects using three different configurations.

\begin{table}[h!]
  \centering
\captionsetup{font=footnotesize,labelsep=period}  
  \caption{Summary of the objects used for grasping and slip detection tests, and their properties including dimension and weight. The dimensions include the objects' radius (for a rectangular object, this refers to half of its width).}
  \newcolumntype{C}[1]{>{\centering\arraybackslash}m{#1}}
  
  \resizebox{0.5\textwidth}{!}{%
    \begin{tabular}{|C{2cm}|C{2cm}|C{2cm}|C{2cm}|C{2cm}|}
      \hline
      Bottle 1 & Bottle 2 & Box 1 & Mango & Box 2 \\
      \hline
      \includegraphics[width=1cm, height=1.5cm]{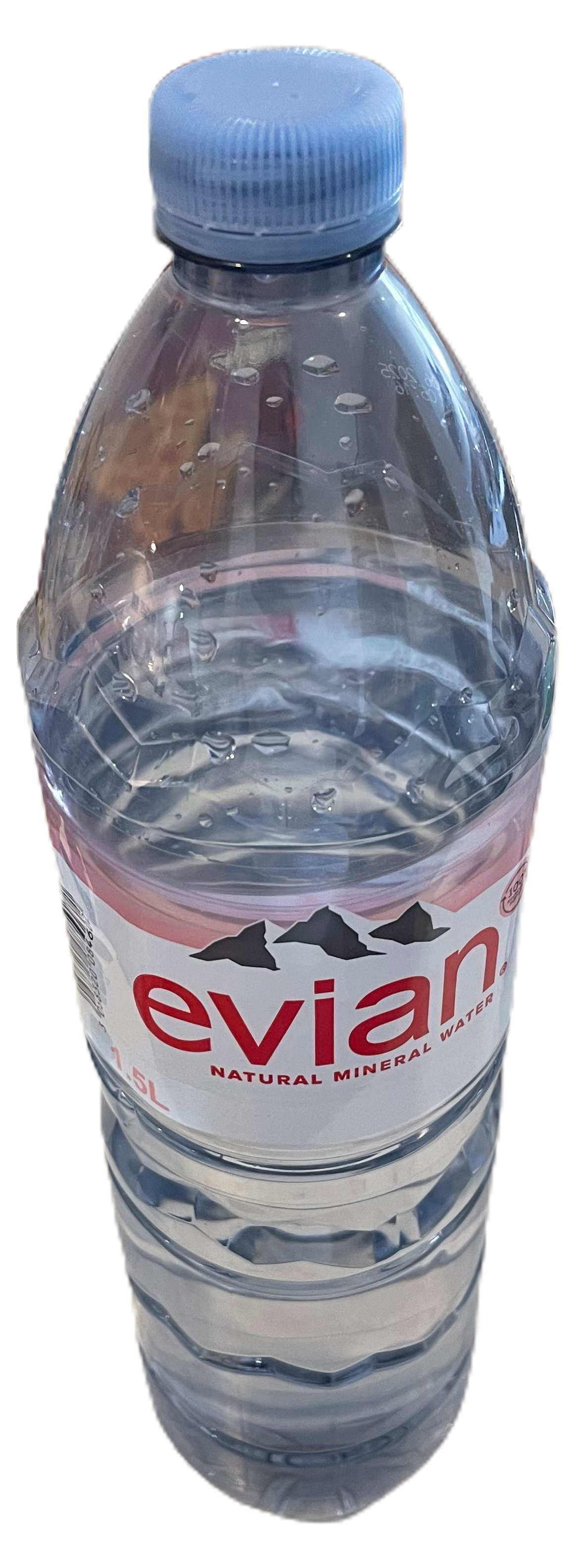} & \includegraphics[width=1cm, height=1.5cm]{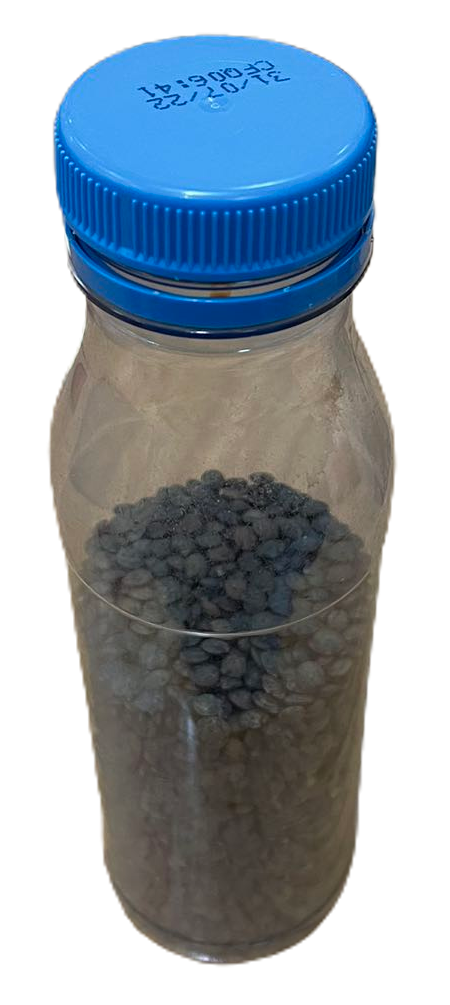} & \includegraphics[width=1.5cm, height=1.5cm]{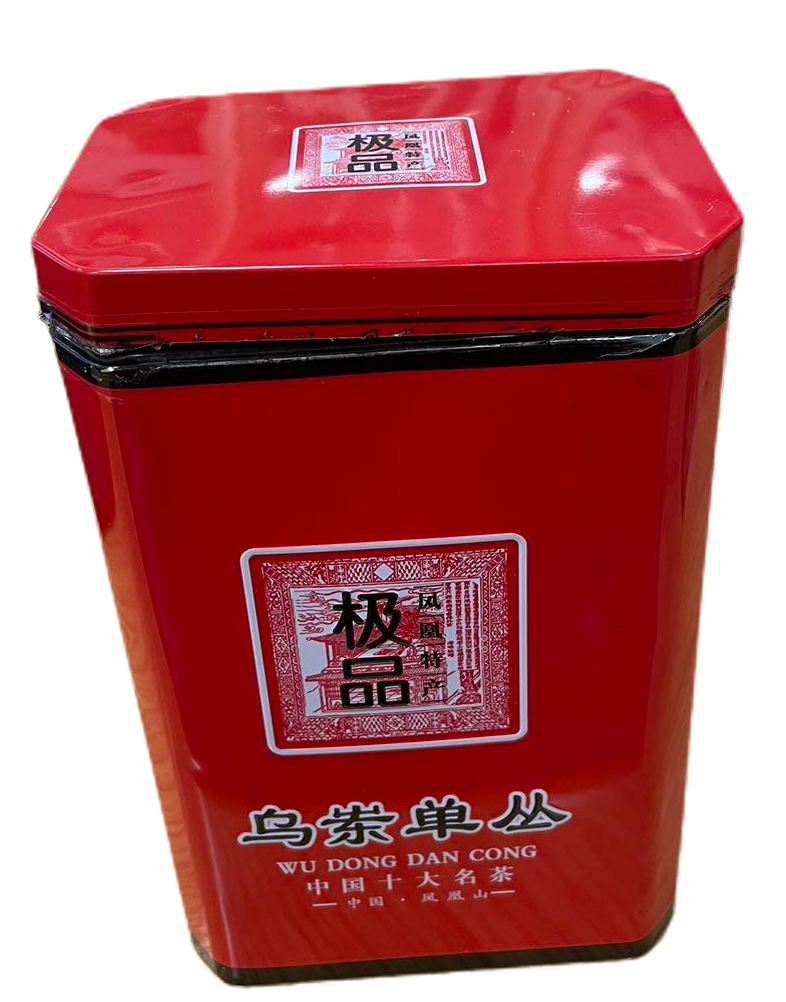} & \includegraphics[width=1.5cm, height=1.5cm]{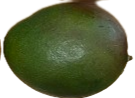} & \includegraphics[width=1.5cm, height=1.5cm]{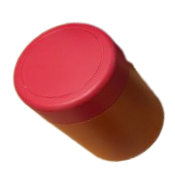} \\
      \hline
      42 mm & 20 mm & 30 mm & 35 mm & 26 mm \\
      \hline
      1015 g & 400 g & 380 g & 300 g & 260 g \\
      \hline
    \end{tabular}
  } 
  
  \vspace{0.5em} 
  
  \resizebox{0.5\textwidth}{!}{%
    \begin{tabular}{|C{2cm}|C{2cm}|C{2cm}|C{2cm}|C{2cm}|}
      \hline
      Box 3 & Apple & Orange & Peper & Glasses \\
      \hline
      \includegraphics[width=1.5cm, height=1.5cm]{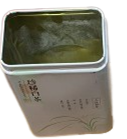} & \includegraphics[width=1.5cm, height=1.5cm]{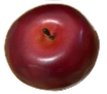} & \includegraphics[width=1.5cm, height=1.5cm]{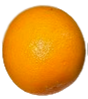} & \includegraphics[width=1.5cm, height=1.5cm]{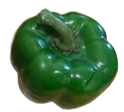} & \includegraphics[width=1.5cm, height=1.5cm]{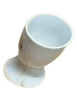} \\
      \hline
      28 mm & 35 mm & 33 mm & 35 mm & 10 mm \\
      \hline
      130 g & 94 g & 82 g & 74 g & 63 g \\
      \hline
    \end{tabular}
  } 
  
  \vspace{0.5em} 
  
  \resizebox{0.5\textwidth}{!}{%
    \begin{tabular}{|C{2cm}|C{2cm}|C{2cm}|C{2cm}|C{2cm}|}
      \hline
      Bottle 3 & potato & tomato & block & corn \\
      \hline
      \includegraphics[width=1.5cm, height=1.5cm]{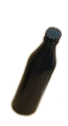} & \includegraphics[width=1.5cm, height=1.5cm]{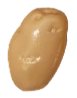} & \includegraphics[width=1.5cm, height=1.5cm]{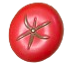} & \includegraphics[width=1.5cm, height=1.5cm]{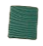} & \includegraphics[width=1.5cm, height=1.5cm]{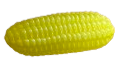} \\
      \hline
      5 mm &  7 mm & 10 mm & 7 mm & 4 mm \\
      \hline
      16 g & 16 g & 14 g & 14 g & 6 g \\
      \hline
    \end{tabular}
  } 
  \label{ex}
\end{table}

\begin{figure*}[t]
    \centering
\captionsetup{font=footnotesize,labelsep=period}      \includegraphics[width=0.95\textwidth,height=2.5 in]{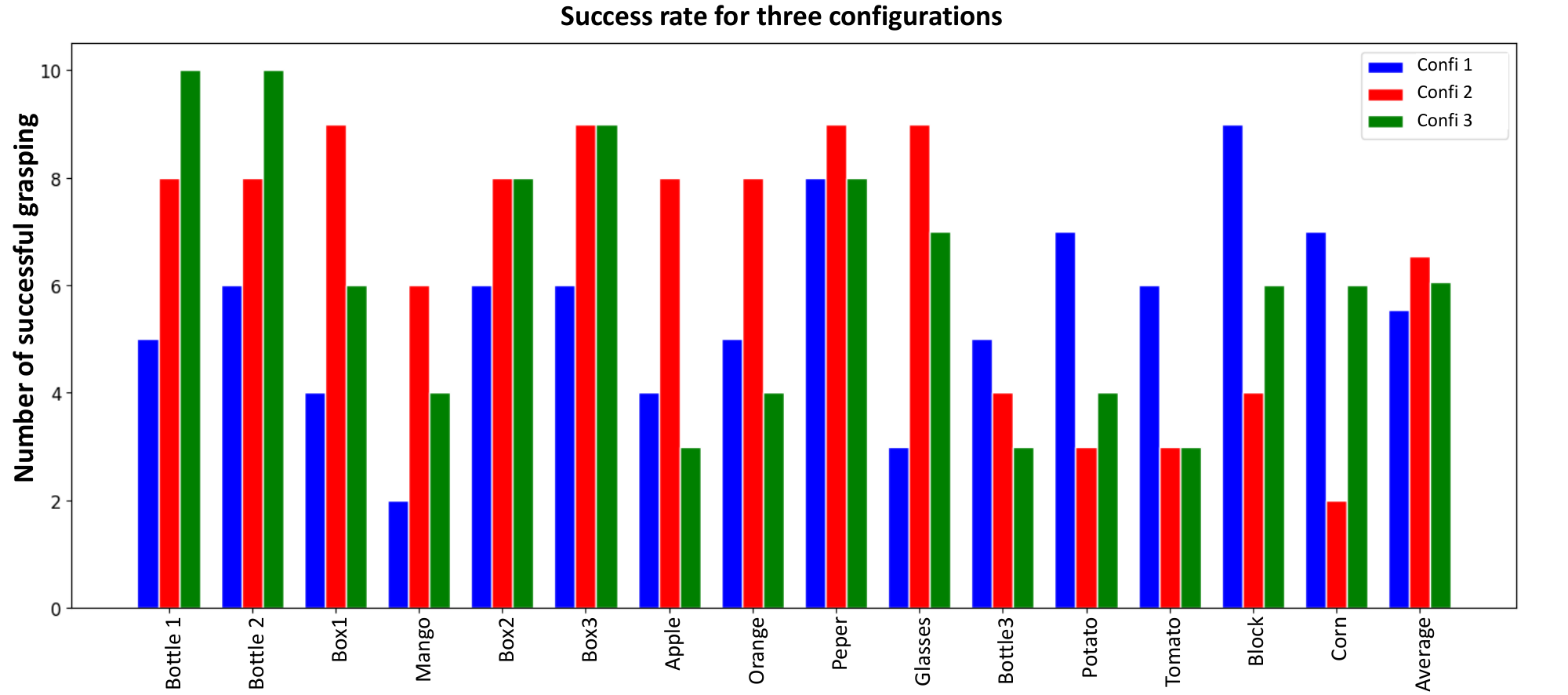}
    \caption{Success rate of grasping various objects using different gripper configurations.}
    \label{Results}
    \vspace{-3mm} 
\end{figure*}

The results of the experiment are shown in Fig. \ref{Results}, where the different colors of bars represent different configurations. The overall success rate is approximately 80\%, thanks to the gripper's re-configurable capability, allowing various configurations to leverage their unique strengths. Smaller objects like the pizza model and small bottles posed challenges due to their small dimensions. As indicated by the prior analysis of the workspace, the gripper encounters difficulties with objects that are either too small (less than 10mm in radius) or too large (exceeding 125mm in radius).

Configuration 1 excels in the manipulation of smaller objects but lags behind in comparison to the others. Configuration 2 demonstrates the best overall performance and has a high success rate in terms of grasping large and medium-sized objects. In contrast, Configuration 3 particularly excels in grasping larger, elongated objects. In conclusion, the gripper's morphological differences enabled by reconfigurable design significantly enhance its grasping capabilities. This adaptability is crucial in real-world scenarios where a diversity of objects demands varied manipulation strategies.

\subsection{Performance Evaluation for Different Algorithms}
In this set of experiments, our main goal is to assess the effectiveness of two slip detection methods used for ensuring a secure grasp on objects. The evaluation is based on three key metrics:
\begin{itemize}
\item Success Rate: This metric evaluates how effectively the slip detection methods prevent object slippage during the grasp. A higher success rate indicates better performance in avoiding unintended slippage.
\item Time Cost: This metric measures the speed at which the gripper identifies slip events and reacts appropriately. Quick detection and response are crucial for efficient and timely robotic operations.
\item Contact Force: This metric examines the amount of force applied during grasping. It's crucial to maintain a secure grip without applying excessive force, especially when manipulating delicate and soft objects.
\end{itemize}

\begin{table}[t]
\centering
\captionsetup{font=footnotesize,labelsep=period} 
\caption{Comparative analysis between the threshold-based method and the GNN-based method for slip detection.}
\begin{tabular}{lcc}
\toprule
                   & \textbf{Threshold Approach} & \textbf{GNN Approach}   \\ 
\midrule
Iteration          & 97.5      & 30    \\
Contact Force (N)          & 3.2       & 2.35  \\
Success rate (\%)  & 80        & 80    \\
\bottomrule
\end{tabular}
\label{your-label-here2}
\vspace{-2mm}  
\end{table}

Based on quantitative data, the GNN-based method demonstrates remarkable efficiency, surpassing the threshold-based method (baseline) by a factor of three.
In terms of force control, the GNN-based method exhibits a substantial 36\% improvement over the baseline, which indicates its superior capability in adjusting forces during manipulation tasks. However, the GNN-based method occasionally misinterprets insufficient gripping forces as adequate. We will investigate the development of high-resolution tactile skin in the future.

\subsection{Evaluation Of In-Hand Manipulation}

\begin{figure}[t]
            \centering
\captionsetup{font=footnotesize,labelsep=period}         \includegraphics[width=0.5\textwidth,height=1.0 in]{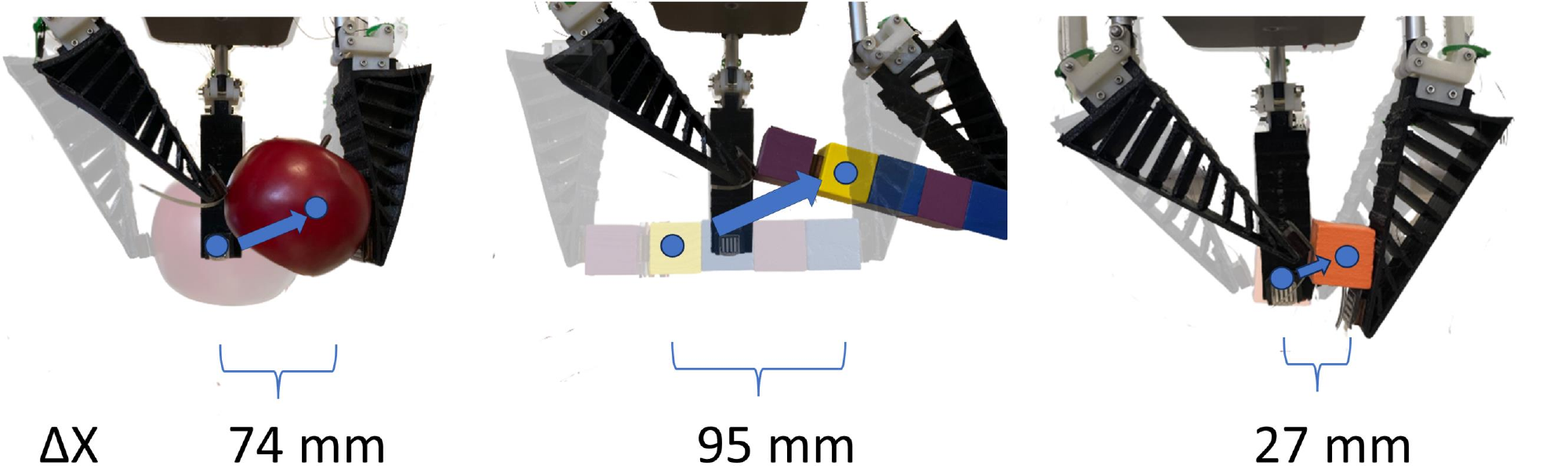}
              \caption{Comparisons between the range of in-hand manipulation between the simulation from the workspace analysis and the physical experimental results for manipulating objects with various sizes and shapes. $\Delta{X}$ indicates the maximum in-hand object translation distance. }
              \label{Range}
 \vspace{-2mm}              
\end{figure}

In this final experiment, we compare the simulation results and experimental results for in-hand manipulation. We chose three objects as examples, as shown in Fig. \ref{Range}. Based on our experiments and previous workspace analysis, we discovered that actual displacements in the physical environment align with our simulation. However, real-world displacements were slightly smaller than our simulated results from workspace analysis. This discrepancy arises since our simulations are based on ideal conditions, assuming perfect contact between the robotic fingers and the object, with guaranteed successful grasping. In reality, factors such as the object's shape, weight, and surface material can influence simulation results. We will enhance the precision of simulation in the future.

\begin{table}[t]
\centering
\captionsetup{font=footnotesize,labelsep=period}
\caption{Comparison between Maximum movement for manipulating different objects in real-world and simulation.}
\begin{tabular}{ccc}
\toprule
\textbf{Object Radius} & \textbf{Real-World Movement} & \textbf{Simulation Movement} \\
\midrule
35 mm & 74 mm & 90 mm\\
65 mm & 95 mm & 100 mm\\
12 mm & 27 mm & 35 mm  \\
\bottomrule
\end{tabular}
\label{table2}
\end{table}

\begin{figure}[t]
    \centering
\captionsetup{font=footnotesize,labelsep=period}    
\includegraphics[width=0.5\textwidth,height=1.5 in]{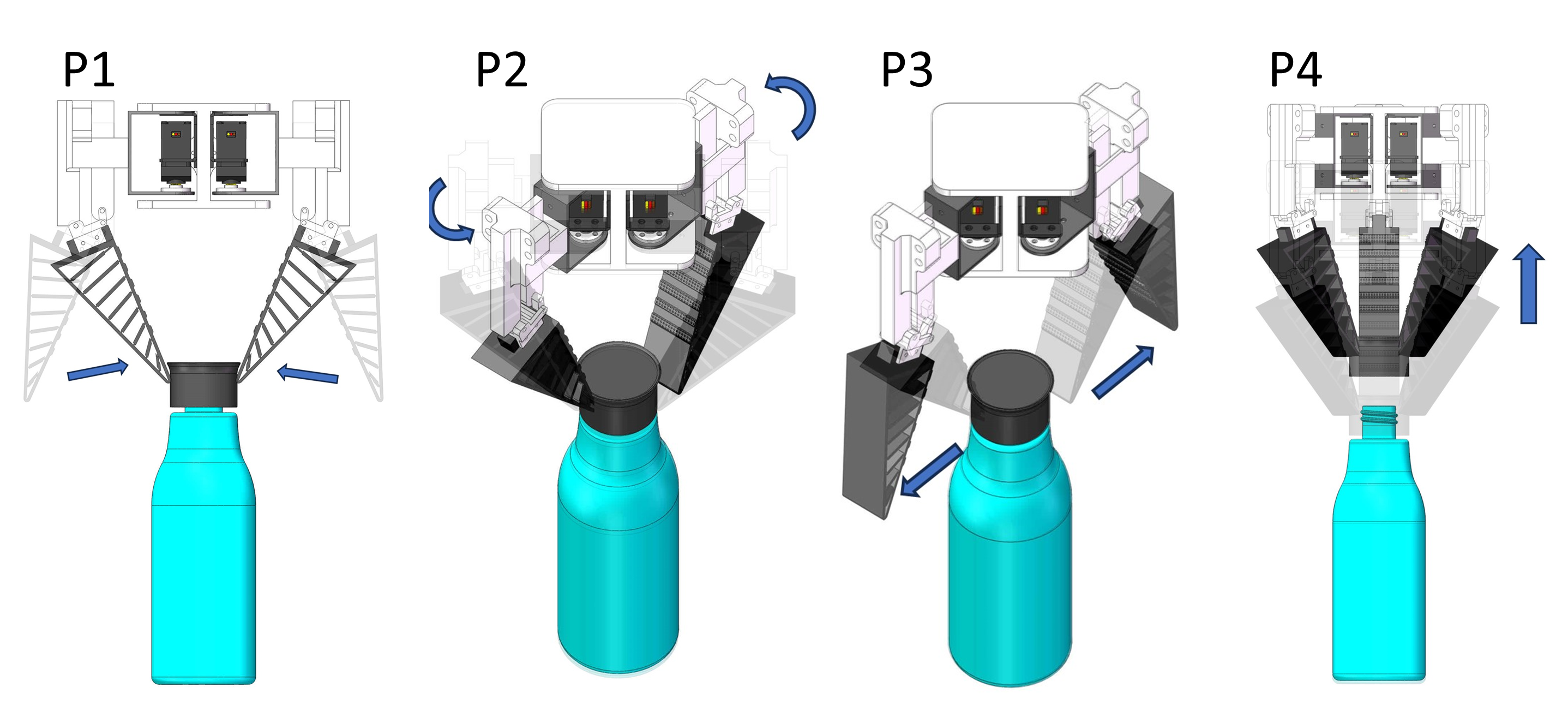}
    \caption{Illustration of the procedures (P1-P4) for the bottle cap removal task.}
    \label{fig:enter-label}
    \vspace{-1mm}
\end{figure}
To demonstrate the effective in-hand manipulation capability of the TacRF-Gripper, we assembled the gripper on a Franka robotic arm and conducted a bottle cap removal task.  The whole task can be segmented into four procedures (see Fig. \ref{fig:enter-label}, P1-P4):
\begin{itemize}
\item Adjust Gripper Pose and Grasp (P1): This step involves determining the optimal angle for the robotic gripper's interaction with the bottle cap. 
\item Bottle Cap Rotation (P2): The robotic gripper rotates the bottle cap using a servomotor located at the base of its fingers while maintaining consistent contact.
\item Reorientation and Iteration (P3): This procedure involves repositioning the pose of the robotic gripper and then repeating Procedures P1 and P2 iteratively until the cap becomes loose.
\item Bottle Cap Lift-off (P4): In this final step, the robotic gripper grasps the cap with all three fingers and performs a lift-off action using the robot arm.
\end{itemize}

To assess the performance of bottle cap removal, experiments are conducted five times, with data collected specifically during `Cap Rotation'. Whenever a slip is detected, the servomotor stops to allow for re-grasping. Therefore, the iteration metric serves as an indicator of both the algorithm's efficiency and robustness in handling the task. 

We compare the GNN-based and threshold-based approach for this application. The results are summarized in Table \ref{your-label-here2}. According to the experimental results, the GNN-based method outperformed the threshold-based method (baseline) in maintaining grip force during bottle cap-twisting tasks. The GNN-based approach leads to reduced task durations, reduced contact force, and increased success rate.

\begin{table}[t]
\centering
\captionsetup{font=footnotesize,labelsep=period} 
\caption{Comparisons between the threshold-based and GNN-based approaches for the  bottle cap removal task.}
\begin{tabular}{lcc}
\toprule
                   & \textbf{Threshold Approach} & \textbf{GNN Approach}\\
\midrule
Iteration   & 1060    & 780  \\
Contact Force (N)  & 5.5    & 4.3  \\
Success rate (\%)  & 72    & 88\\
\bottomrule
\end{tabular}
\label{your-label-here4}
\vspace{-6mm} 
\end{table}

\section{Discussion and Conclusion  }

In this paper, we introduce the TacFR-Gripper, a reconfigurable Fin Ray-based soft tactile gripper capable of grasping and manipulating objects of various sizes. We incorporated a tactile sensor array onto the soft Fin-Ray finger, which can be known as the tactile skin. We utilize a GNN-based algorithm for tactile data processing, which significantly enhances slip detection capabilities. Through rigorous testing and comprehensive analysis, we demonstrate the TacFR-Gripper's proficiency in handling a wide range of delicate objects, and showing its potential applicability in a range of fields, including industrial automation and medical assistance. Moreover, experimental comparisons indicate that the proposed GNN-based tactile perception approach outperforms threshold-based control methods in terms of accuracy and efficiency.

In our future work, we plan to integrate tactile skin across the entire gripper surface.  More complex tasks will be carried out to further evaluate the capabilities of the TacFR-Gripper for contact-rich dexterous manipulation. 


\bibliographystyle{IEEEtran}

\bibliography{IEEEabrv,ref}

\end{document}